\documentclass[11pt,a4paper]{article}
\usepackage[hyperref]{acl2021}
\usepackage{times}
\usepackage{latexsym}
\usepackage{color}
\usepackage{pifont}
\usepackage{makecell}

\usepackage{stackengine}
\newcommand\xrowht[2][0]{\addstackgap[.5\dimexpr#2\relax]{\vphantom{#1}}}
\usepackage{latexsym}
\usepackage{amsthm,amsmath,amssymb}
\usepackage{multirow}
\usepackage{array}
\usepackage{mathrsfs}
\usepackage{graphicx}
\usepackage{floatrow}
\usepackage{booktabs}
\usepackage{algorithm}
\usepackage[normalem]{ulem}
\useunder{\uline}{\ul}{}
\usepackage{tikz}
\usepackage{pgfplots}
\usepackage{subcaption}
\usepackage{algorithm}
\usepackage{algorithmicx}
\usepackage{algpseudocode}

\usepackage{microtype}

\aclfinalcopy 

\newcommand*\samethanks[1][\value{footnote}]{\footnotemark[#1]}

\title{A Unified Generative Framework for Aspect-Based Sentiment Analysis
}

\author{Hang Yan\textsuperscript{1,}\thanks{\ \  Equal contribution.} , Junqi Dai\textsuperscript{1,}\samethanks \ , Tuo Ji\textsuperscript{1}, Xipeng Qiu\textsuperscript{1,2}\thanks{\ \  Corresponding author.}, Zheng Zhang\textsuperscript{3}\\
  \textsuperscript{1}Shanghai Key Laboratory of Intelligent Information Processing, Fudan University \\
  \textsuperscript{1}School of Computer Science, Fudan University \\
  \textsuperscript{2}Pazhou Lab, Guangzhou, China \\
  \textsuperscript{3}New York University \\
  \texttt{\{hyan19,jqdai19,tji19,xpqiu\}@fudan.edu.cn}\\
  \texttt{zz@nyu.edu}\\}

\date{}

\begin{document}
\maketitle
\begin{abstract}
  Aspect-based Sentiment Analysis (ABSA) aims to identify the aspect terms, their corresponding sentiment polarities, and the opinion terms. There exist seven subtasks in ABSA. Most studies only focus on the subsets of these subtasks, which leads to various complicated ABSA models while hard to solve these subtasks in a unified framework. In this paper,  we redefine every subtask target as a sequence mixed by pointer indexes and sentiment class indexes, which converts all ABSA subtasks into a unified generative formulation. Based on the unified formulation, we exploit the pre-training sequence-to-sequence model BART to solve all ABSA subtasks in an end-to-end framework. Extensive experiments on four ABSA datasets for seven subtasks demonstrate that our framework achieves substantial performance gain and provides a real unified end-to-end solution for the whole ABSA subtasks, which could benefit multiple tasks\footnote{Code is available at \url{https://github.com/yhcc/BARTABSA}.}. \end{abstract}

\section{Introduction}
Aspect-based Sentiment Analysis (ABSA) is the fine-grained Sentiment Analysis (SA) task, which aims to identify the aspect term ($a$), its corresponding sentiment polarity ($s$),   and the opinion term ($o$).
For example, in the sentence ``\textit{The \textcolor[rgb]{1.00,0.00,0.00}{drinks} are always \textcolor[RGB]{101,139,243}{well made} and \textcolor[rgb]{1.00,0.00,0.00}{wine selection} is \textcolor[RGB]{101,139,243}{fairly priced}}'', the aspect terms are ``\textit{\textcolor[rgb]{1.00,0.00,0.00}{drinks}}'' and ``\textit{\textcolor[rgb]{1.00,0.00,0.00}{wine selection}}'', and their sentiment polarities  are both ``\textcolor[RGB]{0, 176, 80}{positive}'', and the opinion terms are  ``\textit{\textcolor[RGB]{101,139,243}{well made}}'' and ``\textit{\textcolor[RGB]{101,139,243}{fairly priced}}''.
\begin{figure}[t]
  \includegraphics[width=1\textwidth]{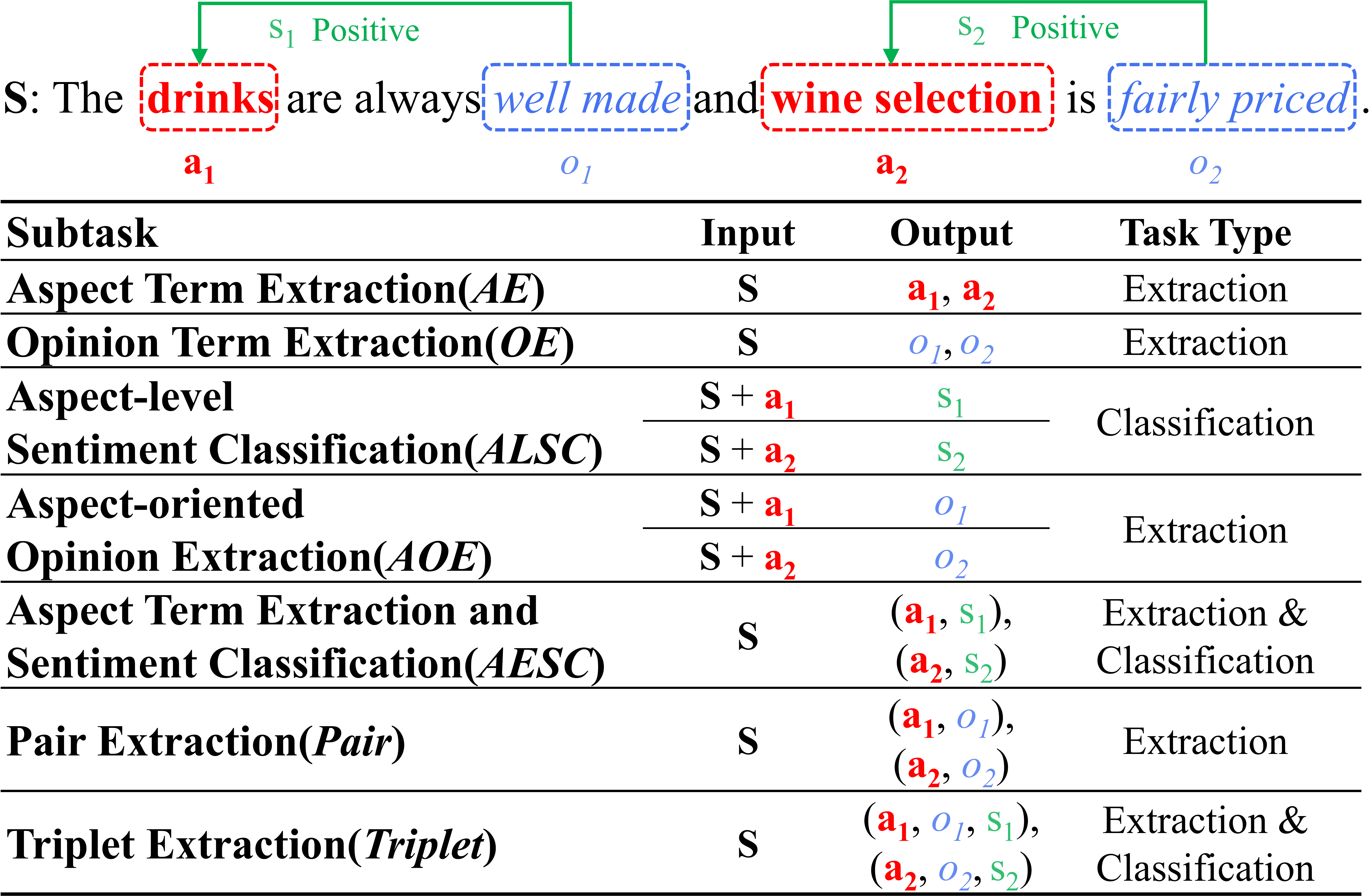}
  \caption{Illustration of seven ABSA subtasks.
  }\label{fig:sent_example}
\end{figure}
Based on the combination of the $a$, $s$, $o$, there exist seven subtasks in ABSA. We summarize these subtasks in Figure~\ref{fig:sent_example}. Specifically, their definitions are  as follows:

$\bullet$ Aspect Term Extraction(\emph{AE}): Extracting all the aspect terms from a sentence.

$\bullet$ Opinion Term Extraction (\emph{OE}): Extracting all the opinion terms from a sentence.

$\bullet$ Aspect-level Sentiment Classification (\emph{ALSC}): Predicting the sentiment polarities for every given aspect terms in a sentence.

$\bullet$ Aspect-oriented Opinion Extraction (\emph{AOE}): Extracting the paired opinion terms for every given aspect terms in a sentence.

$\bullet$ Aspect Term Extraction and  Sentiment Classification (\emph{AESC}): Extracting the aspect terms as well as the corresponding sentiment polarities simultaneously.

$\bullet$ Pair Extraction (\emph{Pair}):  Extracting the aspect terms as well as the corresponding opinion terms simultaneously.

$\bullet$ Triplet Extraction (\emph{Triplet}): Extracting all aspects terms with their corresponding opinion terms and sentiment polarity simultaneously.

Although these ABSA subtasks are strongly related, most of the existing work only focus 1$\sim$3 subtasks individually. The following divergences make it difficult to solve all subtasks in a unified framework.
\begin{enumerate}
\setlength{\itemsep}{1pt}%
\setlength{\parskip}{1pt}%
  \item \textit{Input:} Some subtasks (  \emph{AE}, \emph{OE},  \emph{AESC},  \emph{Pair}  and \emph{Triplet}) only take the text sentence as input, while the remained subtasks ( \emph{ALSC} and \emph{AOE}) take the text and a given aspect term as input.
  \item \textit{Output:}  Some tasks (\emph{AE},  \emph{OE}, \emph{ALSC}, \emph{AOE}) only output a certain type from $a$, $s$ or $o$, while the remained tasks (\emph{AESC}, \emph{Pair} and \emph{Triplet}) return compound output as the combination of $a$, $s$ and $o$.
  \item \textit{Task Type:} There are two kinds of tasks: extraction task (extracting aspect and opinion) and classification task (predicting sentiment).
\end{enumerate}


Because of the above divergences, a myriad of previous  works only focus on the subset of these subtasks. However, the importance of solving the whole ABSA subtasks in a unified framework remains significant. Recently, several works make attempts on this track.  Some methods\citep{DBLP:conf/aaai/PengXBHLS20,DBLP:journals/corr/abs-2101-00816} apply the pipeline model to output the $a$, $s$, $o$ from the inside sub-models separately. However, the pipeline process is  not end-to-end. Another line follows the sequence tagging method by extending the tagging schema~\citep{DBLP:conf/emnlp/XuLLB20}. However, the  compositionality of candidate labels hinders the performance. In conclusion, the existing methods can hardly solve all the subtasks by a unified framework  without relying on the sub-models or changing the model structure  to adapt to all ABSA subtasks.

Motivated by the above observations, we propose a unified generative framework to address all the ABSA subtasks. We first formulate all these subtasks as a  generative task, which could handle the obstacles on the input, output, and task type sides and adapt to all the subtasks without any model structure changes. Specifically, we model the extraction and classification tasks as the pointer indexes and class indexes generation, respectively.
Based on the unified task formulation, we use the sequence-to-sequence pre-trained model BART~\citep{DBLP:conf/acl/LewisLGGMLSZ20} as our backbone to generate the target sequence in an end-to-end process.  To validate the effectiveness of our method, we conduct extensive experiments on public datasets. The comparison results demonstrate that our proposed framework outperforms most state-of-the-art (SOTA) models in every subtask.

In summary, our main contributions are as follows:

$\bullet$ We formulate both the extraction task and classification task of ABSA into a unified index generation problem. Unlike previous unified models, our method needs not to design specific decoders for different output types.

$\bullet$ With our re-formulation, all ABSA subtasks can be solved in sequence-to-sequence framework, which is easy-to-implement and can be built on the pre-trained models, such as BART.

$\bullet$ We conduct extensive experiments on  four public datasets, and each dataset contains a subset of all ABSA subtasks. To the best of our knowledge, it is the first work to evaluate a model on all ABSA tasks.

$\bullet$ 
The experimental results show that our proposed framework significantly outperforms recent SOTA methods.

\section{Background}
\subsection{ABSA Subtasks}
In this section, we first review the existing studies on single output subtasks, and then turn to studies focusing on the compound output subtasks.

\subsubsection{Single Output Subtasks}
Some researches mainly focus on the single output subtasks. The \emph{AE}, \emph{OE}, \emph{ALSC} and \emph{AOE} subtasks only output one certain type from $a$, $s$ or $o$.

\textbf{\emph{AE}} Most studies treat \emph{AE} subtask as a  sequence tagging problem~\citep{DBLP:conf/emnlp/LiL17,DBLP:conf/acl/XuLSY18,DBLP:conf/ijcai/LiBLLY18}. Recent works explore sequence-to-sequence learning on \emph{AE} subtask, which obtain promissing results  especially with the pre-training language models~\citep{dblp:conf/acl/malwxw19, DBLP:conf/acl/LiCQLS20}.

\textbf{\emph{OE}} Most studies treat \emph{OE} subtask  as an auxiliary task~\citep{DBLP:conf/emnlp/WangPDX16,DBLP:conf/aaai/WangPDX17, DBLP:conf/acl/PanW18, DBLP:conf/acl/ChenQ20,DBLP:conf/acl/HeLND19}. Most works can only extract the unpaired aspect and opinion terms\footnote{It is also referred to as the AE-OE co-Extraction.}. In this case, opinion terms are independent of aspect terms.

\textbf{\emph{ALSC}} \citet{DBLP:conf/coling/TangQFL16} use the long short term memory (LSTM) network to enhance the interactions between aspects and context words. \citet{DBLP:conf/emnlp/WangHZZ16,DBLP:conf/eacl/ZhangL17,DBLP:conf/ijcai/MaLZW17,DBLP:conf/aaai/TayTH18a} incorporate the attention mechanism into the LSTM-based neural network models to model relations of aspects and their contextual words. Other model structures such as convolutional neural network (CNN)~\citep{DBLP:conf/acl/LamLSB18,DBLP:conf/acl/LiX18}, gated neural network ~\citep{DBLP:conf/aaai/ZhangZV16,DBLP:conf/acl/LiX18}, memory neural network ~\citep{DBLP:conf/emnlp/TangQL16,DBLP:conf/emnlp/ChenSBY17} have also been applied.

\textbf{\emph{AOE}} This subtask is first introduced by \citet{DBLP:conf/naacl/FanWDHC19}  and they propose the datasets for this subtask. Most studies apply sequence tagging method for this subtask~\citep{DBLP:conf/aaai/WuZDHC20,DBLP:conf/emnlp/VeysehNDDN20}.
\subsubsection{Compound Output Subtasks}
Some researchers pay more attention and efforts to the subtasks with compound output. We review them as follows:

\textbf{\emph{AESC}}. One line follows pipeline method to solve this problem. Other works utilize unified tagging schema~\citep{DBLP:conf/emnlp/MitchellAWD13,DBLP:conf/emnlp/ZhangZV15,DBLP:conf/aaai/LiBLL19} or multi-task learning~\citep{DBLP:conf/acl/HeLND19,DBLP:conf/acl/ChenQ20} to avoid the error-propagation problem~\citep{DBLP:conf/emnlp/MaLW18}. Span-based \emph{AESC}  works are also proposed recently~\citep{DBLP:conf/acl/HuPHLL19}, which can tackle the sentiment inconsistency problem in the unified tagging schema.

\textbf{\emph{Pairs}} \citet{DBLP:conf/acl/ZhaoHZLX20}  propose to extract all ($a$, $o$)  pair-wise relations from scratch. They propose a multi-task learning framework based on the span-based extraction method  to handle this subtask.

\textbf{\emph{Triplet}} This subtask is proposed by \citet{DBLP:conf/aaai/PengXBHLS20}  and gains increasing interests recently. \citet{DBLP:conf/emnlp/XuLLB20} design the position-aware tagging schema and apply model based on CRF~\citep{DBLP:conf/icml/LaffertyMP01} and Semi-Markov CRF~\citep{DBLP:conf/nips/SarawagiC04}. However, the time complexity limits the model to detect the aspect term with  long-distance opinion terms.  \citet{DBLP:journals/corr/abs-2101-00816} formulate \emph{Triplet} as a two-step MRC problem, which applies the pipeline method.

\begin{figure*}[t!]
  \includegraphics[width=1\textwidth]{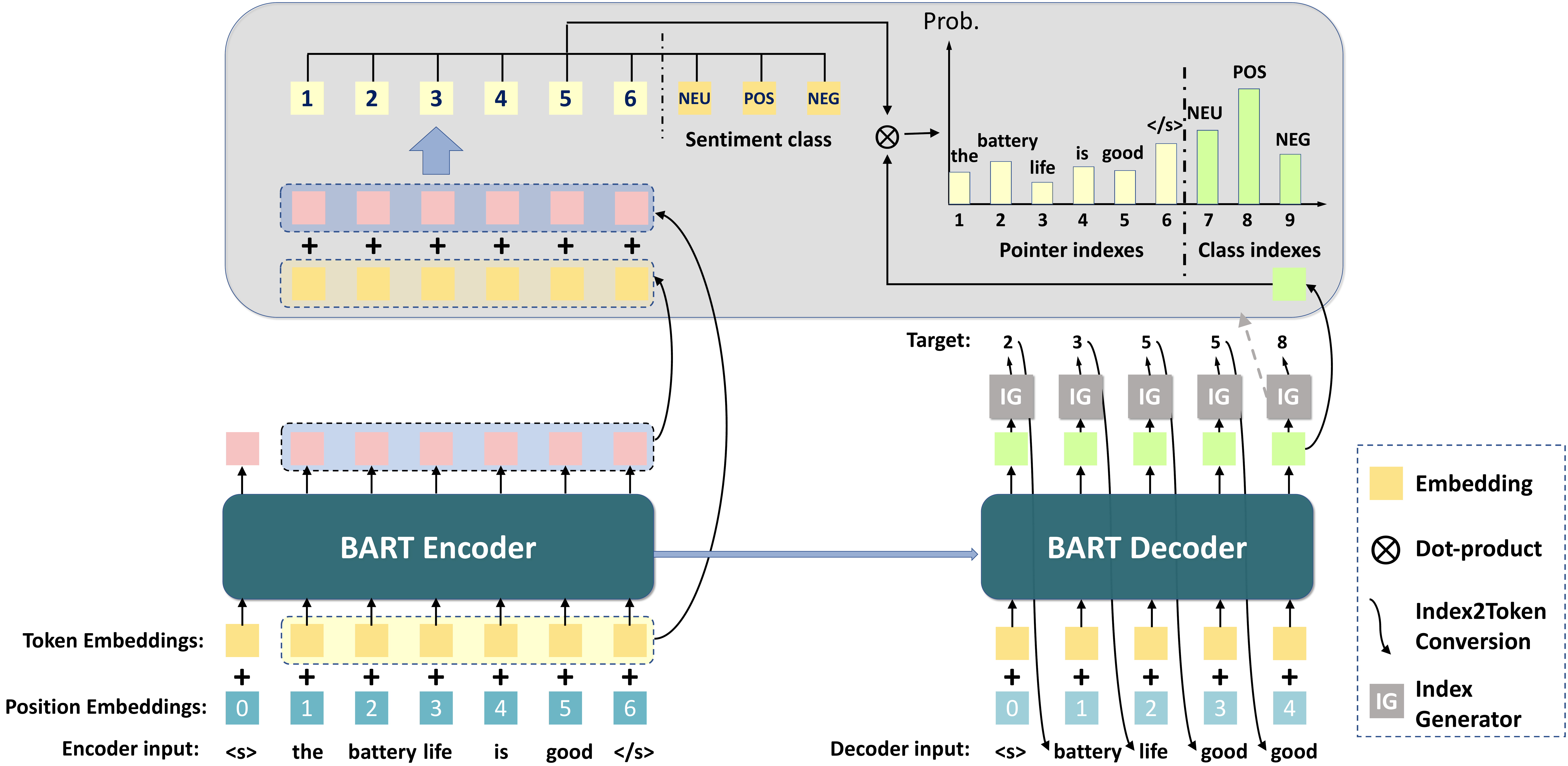}
  \caption{Overall architecture of the framework. This shows an example generation process for the \emph{Triplet} subtask where the source is ``\emph{\textless{}s\textgreater the battery life is good \textless{}/s\textgreater}'' and the target is ``\emph{2 3 5 5 8 6}''(Only partial decoder sequence is shown where the 6 (\textless{}/s\textgreater) should be the next  generation index). The ``Index2Token Conversion'' converts the index to tokens. Specifically, the pointer index will be converted to its corresponding token in the source text, and the class index will be converted to corresponding class tokens. Embedding vectors in \colorbox[RGB]{252,227,138}{\color[RGB]{252,227,138}{ll}} boxes are retrieved from same embedding matrix. We use different position embeddings in the source and target for better generation performance. }
  \label{fig:model}
\end{figure*}

\begin{figure}[h]
  \includegraphics[width=1\textwidth]{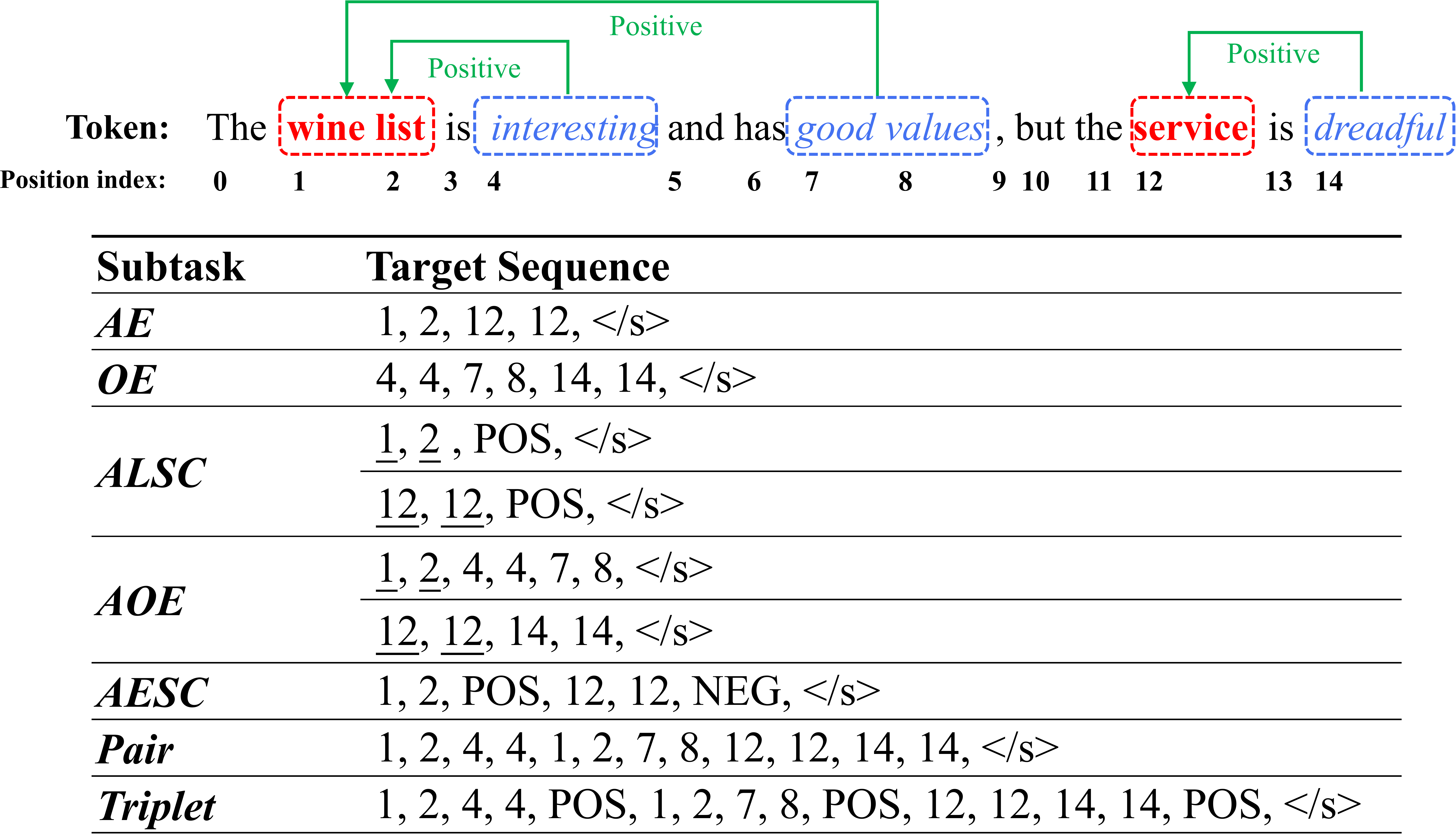}
  \caption{Target sequences for different subtasks. The underlined indexes are given in advance. We convert the sentiment class index to the corresponding class token for better understanding.}\label{fig:detail}
\end{figure}

\subsection{Sequence-to-Sequence Models}
The sequence-to-sequence framework has been long studied in the NLP field to tackle various tasks \cite{DBLP:conf/nips/SutskeverVL14,DBLP:conf/emnlp/ChoMGBBSB14,DBLP:conf/nips/VinyalsFJ15,DBLP:conf/emnlp/LuongPM15}. Inspired by the success of PTMs (pre-trained models) \cite{DBLP:journals/corr/abs-2003-08271,DBLP:conf/naacl/PetersNIGCLZ18,dblp:conf/naacl/devlinclt19,brown2020language}, \citet{DBLP:conf/icml/SongTQLL19,DBLP:journals/jmlr/RaffelSRLNMZLL20,DBLP:conf/acl/LewisLGGMLSZ20} try to pre-train sequence-to-sequence models. Among them, we use the BART \cite{DBLP:conf/acl/LewisLGGMLSZ20} as our backbone, while the other sequence-to-sequence pre-training models can also be applied in our architecture to use the pointer mechanism \cite{DBLP:conf/nips/VinyalsFJ15}, such as MASS \cite{DBLP:conf/icml/SongTQLL19}.

BART is a strong sequence-to-sequence pre-trained model for Natural Language Generation (NLG). BART is a denoising autoencoder composed of several transformer \cite{dblp:conf/nips/vaswanispujgkp17} encoder and decoder layers. It is worth noting that the BART-Base model contains a 6-layer encoder and 6-layer decoder, which makes it similar number of parameters\footnote{Because of the cross-attention between encoder and decoder, the number of parameters of BART is about 10\% larger than its counterpart of BERT \cite{DBLP:conf/acl/LewisLGGMLSZ20}.} with the BERT-Base model.  BART is pretrained on denoising tasks where the input sentence is noised by some methods, such as masking and permutation.  The encoder takes the noised sentence as input, and the decoder will restore the original sentence in an autoregressive manner.

\section{Methodology}
Although there are two types of tasks among the seven ABSA subtasks, they can be formulated under a generative framework. In this part, we first introduce our sequential representation for each ABSA subtask. Then we detail our method, which utilizes  BART to generate these sequential representations.

\subsection{Task Formulation}
As depicted in Figure~\ref{fig:sent_example}, there are two types of tasks, namely the extraction and classification, whose target can be represented as a sequence of pointer indexes and class indexes, respectively. Therefore, we can formulate these two types of tasks in a unified generative framework. We use  $a$, $s$, $o$, to represent the  aspect term, sentiment polarity,and opinion term, respectively. Moreover, we use the superscript $^s$ and $^e$ to denote the start index and end index of a term. For example, $o^s, a^e$ represent the start index of an opinion term $o$ and the end index of an aspect term $a$. We use the $s^p$ to denote the index of sentiment polarity class. The target sequence for each subtask is as follows:

$\bullet$ \emph{AE} : $Y=[a_1^s, a_1^e, ..., a_i^s, a_i^e, ...]$,

$\bullet$ \emph{OE} : $Y=[o_1^s, o_1^e, ..., o_i^s, o_i^e, ...]$,

$\bullet$ \emph{AESC} : $Y=[a_1^s, a_1^e, s^p_1, ..., a_i^s, a_i^e, s^p_i, ...]$,

$\bullet$  \emph{Pair}: $Y=[a_1^s, a_1^e, o_1^s, o_1^e, ..., a_i^s, a_i^e, o_i^s, o_i^e,$$...]$,

$\bullet$ \emph{Triplet} : $Y=[a_1^s, a_1^e, o_1^s, o_1^e, s^p_1, ..., a_i^s, a_i^e, o_i^s, $ $o_i^e, s^p_i, ...]$,

The above subtasks only rely on the input sentence, while for the \emph{ALSC} and \emph{AOE}  subtasks, they also depend on a specific aspect term $a$. Instead of putting the aspect term on the input side, we put them on the target side so that the target sequences are as follows:

$\bullet$ \emph{ALSC} : $Y=[\underline{a^s}, \underline{a^e}, s^p]$,

$\bullet$ \emph{AOE} : $Y=[\underline{a^s}, \underline{a^e}, o_1^s, o_1^e, ..., o_i^s, o_i^e, ...]$,\\
where the underlined tokens are given during inference. Detailed target sequence examples for each subtask are presented in Figure~\ref{fig:detail}.

\begin{table*}[h]
  \centering\small
  \setlength{\tabcolsep}{0.4pt}
  \renewcommand{\arraystretch}{1.2}
  \begin{tabular}{b{0.75cm}b{0.75cm}|b{0.7cm}<{\centering}b{0.7cm}<{\centering}b{0.7cm}<{\centering}b{0.7cm}<{\centering}|b{0.7cm}<{\centering}b{0.7cm}<{\centering}b{0.7cm}<{\centering}b{0.8cm}<{\centering}|b{0.7cm}<{\centering}b{0.7cm}<{\centering}b{0.7cm}<{\centering}b{0.7cm}<{\centering}|b{0.7cm}<{\centering}b{0.7cm}<{\centering}b{0.3cm}<{\centering}b{0.8cm}<{\centering}|b{2.9cm}<{\centering}}
    \Xhline{0.08em}
  \multicolumn{2}{l|}{\multirow{2}{*}{Dataset}} & \multicolumn{4}{c}{14res} & \multicolumn{4}{c}{14lap} & \multicolumn{4}{c}{15res} & \multicolumn{4}{c|}{16res} & \multirow{2}{*}{Subtasks}                                                                                                 \\
    \cline{3-18}
    \multicolumn{2}{l|}{}                         & \#$s$  & \#$a$  & \#$o$  & \#$p$  & \#$s$  & \#$a$  & \#$o$  & \#$p$  & \#$s$  & \#$a$  & \#$o$  & \#$p$  & \#$s$   & \#$a$  & \#$o$ & \#$p$  &                                                                                                                           \\
    \cline{1-19}
    \multirow{2}{*}{\emph{$\mathcal{D}_{17}$}}         & train        & 3044 & 3699 & 3484 & -    & 3048 & 2373 & 2504 & -    & 1315 & 1199 & 1210 & -    & -     & -    & -   & -    & \multicolumn{1}{l}{\multirow{2}{*}{\begin{tabular}[c]{@{}l@{}} \begin{footnotesize}\emph{AE}, \emph{OE}, \emph{ALSC},\end{footnotesize} \\ \begin{footnotesize}\emph{AESC}\end{footnotesize}\end{tabular}}}                                                        \\
                                & test         & 800  & 1134 & 1008 & -    & 800  & 654  & 674  & -    & 685  & 542  & 510  & -    & -     & -    & -   & -    & \multicolumn{1}{l}{}                                                                                                                          \\
    \cline{1-19}
    \multirow{2}{*}{\emph{$\mathcal{D}_{19}$}}          & train        & 1627 & 2643 & -    & -    & 1158 & 1634 & -    & -    & 754  & 1076 & -    & -    & 1079  & 1512 & -   & -    & \multicolumn{1}{l}{\multirow{2}{*}{\footnotesize \emph{AOE} }}                                                                                                      \\
                                & test         & 500  & 865  & -    & -    & 343  & 482  & -    & -    & 325  & 436  & -    & -    & 329   & 457  & -   & -    & \multicolumn{1}{l}{}                                                                                                                           \\
    \cline{1-19}
    \multirow{3}{*}{\emph{$\mathcal{D}_{20a}$}}        & train        & 1300 & -    & -    & 2145 & 920  & -    & -    & 1265 & 593  & -    & -    & 923  & 842   & -    & -   & 1289 & \multirow{3}{*}{\begin{tabular}[l]{@{}l@{}}\begin{footnotesize}\emph{AE}, \emph{OE}, \emph{ALSC}, \emph{AOE}, \end{footnotesize}\\\begin{footnotesize}\emph{AESC}, \emph{Pair},\end{footnotesize} \begin{footnotesize}\emph{Triplet}\end{footnotesize}\end{tabular}} \\
                                & dev          & 323  & -    & -    & 524  & 228  & -    & -    & 337  & 148  & -    & -    & 238  & 210   & -    & -   & 316  &                                                                                                                           \\
                                & test         & 496  & -    & -    & 862  & 339  & -    & -    & 490  & 318  & -    & -    & 455  & 320   & -    & -   & 465  &                                                                                                                           \\
    \cline{1-19}
    \multirow{3}{*}{\emph{$\mathcal{D}_{20b}$}}        & train        & 1266 & -    & -    & 2338 & 906  & -    & -    & 1460 & 605  & -    & -    & 1013 & 857   & -    & -   & 1394 & \multirow{3}{*}{\begin{tabular}[c]{@{}l@{}}\begin{footnotesize}\emph{AE}, \emph{OE}, \emph{ALSC}, \emph{AOE}, \end{footnotesize}\\\begin{footnotesize}\emph{AESC}, \emph{Pair},\end{footnotesize} \begin{footnotesize}\emph{Triplet}\end{footnotesize}\end{tabular}} \\
                                & dev          & 310  & -    & -    & 577  & 219  & -    & -    & 346  & 148  & -    & -    & 249  & 210   & -    & -   & 339  &                                                                                                                           \\
                                & test         & 492  & -    & -    & 994  & 328  & -    & -    & 543  & 148  & -    & -    & 485  & 326   & -    & -   & 514  &                                                                                                                          \\
    \Xhline{0.08em}

  \end{tabular}
  \caption{The statistics of four datasets, where the \#$s$, \#$a$, \#$o$, \#$p$ denote the numbers of sentences, aspect terms, opinion terms, and the \textless{}$a$, $o$\textgreater{}  pairs, respectively. We  use  ``-'' to  denote  the missing  data  statistics of some datasets. The  ``Subtasks'' column refers to the ABSA subtasks that can be applied on the corresponding dataset. }
  \label{tb:data}
\end{table*}

\subsection{Our Model}
As our discussion in the last section, all subtasks can be formulated as taking the $X=[x_1, ..., x_n]$ as input and outputting a target sequence $Y=[y_1, ..., y_m]$, where $y_0$ is the start-of-the-sentence token. Therefore, different ABSA subtasks can be formulated as:
\begin{align}
  P(Y|X) = \prod_{t=1}^{m} P(y_t|X, Y_{<t}).
\end{align}

To get the index probability distribution $P_t = P(y_t|X, Y_{<t})$ for each step, we use a model composed of two components: (1) \textbf{Encoder}; (2) \textbf{Decoder}.

\textbf{Encoder} The encoder part is to encode $X$ into vectors $\mathbf{H}^e$. We use the BART model, therefore, the start of sentence ($<$s$>$) and the end of sentence ($<$/s$>$) tokens will be added to the start and end of $X$, respectively. We ignore the $<$s$>$ token in our equations for simplicity. The encoder part is as follows:
\begin{align}
  \mathbf{H}^e & = \mathrm{BARTEncoder}([x_1, ..., x_n]),
\end{align}
where $\mathbf{H}^e \in \mathbb{R}^{n \times d}$, and $d$ is the hidden dimension.

\textbf{Decoder} The decoder part takes the encoder outputs $\mathbf{H}^e$ and previous decoder outputs $Y_{<t}$ as inputs to get $P_t$. However, the $Y_{<t}$ is  an  index sequence. Therefore, for each $y_t$ in $Y_{<t}$, we first need to use the following $\mathrm{Index2Token}$ module to conduct a conversion
\begin{align}
  \hat{y}_t = \begin{cases}
    X_{y_t},&  \text{if}  \  y_t \  \text{is a pointer index},\\
    C_{y_t - n},&  \text{if} \  y_t \  \text{is a class index},
  \end{cases}
\end{align}
where $C=[c_1, ..., c_l]$ is the class token list\footnote{In our implement, $y_t \in [1, n+l]$. The $x_1$ has the pointer index 1. }.

After that, we use the BART decoder to get the last hidden state
\begin{align}
  \mathbf{h}_t^d & = \mathrm{BARTDecoder}(\mathbf{H}^e; \hat{Y}_{<t}),
\end{align}
where $\mathbf{h}_t^d \in \mathbb{R}^d$. With $\mathbf{h}_t^d$, we predict the token probability distribution $P_t$ as follows:
\begin{align}
  \mathbf{E}^e & = \mathrm{BARTTokenEmbed}(X),\\
  \mathbf{\hat{H}}^e & = \mathrm{MLP}(\mathbf{H}^e), \\
  \mathbf{\bar{H}}^e & = \alpha \mathbf{\hat{H}}^e + (1-\alpha) \mathbf{E}^e, \\
  \mathbf{C}^d & = \mathrm{BARTTokenEmbed}(C), \\
  P_t & = \mathrm{Softmax}([\mathbf{\bar{H}^e};\mathbf{C}^d]  \mathbf{h}_t^d),
\end{align}
where $\mathbf{E}^e,\mathbf{H}^e,\mathbf{\hat{H}}^e,\mathbf{\bar{H}}^e \in \mathbb{R}^{n \times d}$;  $\mathbf{C}^d \in \mathbb{R}^{l \times d}$; and $P_t \in \mathbb{R}^{(n + l)}$ is the final distribution on all indexes.

During the training phase, we use the teacher forcing to train our model and the negative log-likelihood to optimize the model. Moreover, during the inference, we use the beam search to get the target sequence $Y$ in an autoregressive manner. After that, we need to use the decoding algorithm to convert this sequence into the term spans and sentiment polarity. We use the \emph{Triplet} task as an example and present the decoding algorithm in Algorithm \ref{al1}, the decoding algorithm for other tasks are much depicted in the Supplementary Material.

\begin{algorithm}[ht]
  \begin{algorithmic}[1]
    \caption{Decoding Algorithm for the \emph{Triplet} Subtask} \label{al1}
    \Require Number of tokens in the input sentence $n$, target sequence $Y=[y_1, ..., y_m]$ and $y_i \in [1, n+|C|]$
    \Ensure  Target span set $L=\{(a_1^s, a_1^e, o_1^s, o_1^e, s_1), ..., (a_i^s, a_i^e, o_i^s, $ $o_i^e, s_i), ...\}$
    \State $L=\{\}, e=[], i=1$
    \While{$i<=m$}
      \State $y_i = Y[i]$
      \If{$y_i>n$}
      \State  $L.add((e, C_{y_i-n}))$
      \State $e=[]$
      \Else
      \State $e.append(y_i)$
      \EndIf
      \State $i+=1$
    \EndWhile
    \State \Return{$L$}
  \end{algorithmic}
\end{algorithm}

\section{Experiments}
\begin{table*}[!h]
  \centering\small
  \setlength{\tabcolsep}{1pt}
  \renewcommand{\arraystretch}{1.2}

  \begin{tabular}{m{2cm}m{1.3cm}<{\centering}m{2.45cm}m{1.35cm}m{2.65cm}<{\centering}m{0.7cm}<{\centering}m{0.7cm}<{\centering}m{0.7cm}<{\centering}m{0.7cm}<{\centering}m{0.85cm}<{\centering}m{0.85cm}<{\centering}m{0.7cm}<{\centering}}
    \toprule
    Baselines            & \multicolumn{1}{c}{E2E} & Task Formulation             & Backbone           &Datasets                         & \multicolumn{1}{c}{\emph{AE}} & \multicolumn{1}{c}{\emph{OE}} & \multicolumn{1}{c}{\emph{ALSC}} & \multicolumn{1}{c}{\emph{AOE}} & \multicolumn{1}{c}{\emph{AESC}} & \multicolumn{1}{c}{ \emph{Pair} } & \multicolumn{1}{c}{\emph{Triplet}} \\
    \midrule
    SPAN-BERT            & -                       & Span.Extraction             & BERT               & \emph{$\mathcal{D}_{17}$}                  & \ding{51}              & -                      & \ding{51}              & -                       & \ding{51}                & -                         & -                            \\
    IMN-BERT             & \ding{51}               & Seq.Tagging                 & BERT               & \emph{$\mathcal{D}_{17}$}                 & \ding{51}              & \ding{51}              & \ding{51}              & -                       & \ding{51}                & -                         & -                            \\
    RACL-BERT            & -                       & Seq.Tagging                 & BERT               & \emph{$\mathcal{D}_{17}$}                 & \ding{51}              & \ding{51}              & \ding{51}              & -                       & \ding{51}                & -                         & -                            \\
    \midrule
    IOG                  & \ding{51}               & Seq.Tagging                 & LSTM               & \emph{$\mathcal{D}_{19}$}                        & -                      & -                      & -                      & \ding{51}               & -                        & -                         & -                            \\
    LOTN                 & \ding{51}               & Seq.Tagging                 & LSTM               & \emph{$\mathcal{D}_{19}$}                        & -                      & -                      & -                      & \ding{51}               & -                        & -                         & -                            \\
    ONG                  & \ding{51}               & Seq.Tagging                 & BERT               & \emph{$\mathcal{D}_{19}$}                        & -                      & -                      & -                      & \ding{51}               & -                        & -                         & -                            \\
    \midrule
    RINANTE+             & -                       & Seq.Tagging                 & LSTM+CRF           & \emph{$\mathcal{D}_{20a}$},\emph{$\mathcal{D}_{20b}$}                        & \ding{51}              & \ding{51}              & \ding{51}              & -                       & \ding{51}                & \ding{51}                 & \ding{51}                    \\
    CMLA+                & -                       & Seq.Tagging                 & Attention          & \emph{$\mathcal{D}_{20a}$},\emph{$\mathcal{D}_{20b}$}                        & \ding{51}              & \ding{51}              & \ding{51}              & -                       & \ding{51}                & \ding{51}                 & \ding{51}                    \\
    Li-unified+          & -                       & Seq.Tagging                 & LSTM               & \emph{$\mathcal{D}_{20a}$},\emph{$\mathcal{D}_{20b}$}                        & \ding{51}              & \ding{51}              & \ding{51}              & -                       & \ding{51}                & \ding{51}                 & \ding{51}                    \\
    Peng-two-stage       & -                       & Seq.Tagging                 & LSTM+GCN           & \emph{$\mathcal{D}_{20a}$},\emph{$\mathcal{D}_{20b}$}                        & \ding{51}              & \ding{51}              & \ding{51}              & -                       & \ding{51}                & \ding{51}                 & \ding{51}                    \\
    JET-BERT             & \ding{51}               & Seq.Tagging                 & BERT               & \emph{$\mathcal{D}_{20a}$},\emph{$\mathcal{D}_{20b}$}                        & \ding{51}              & \ding{51}              & \ding{51}              & -                       & \ding{51}                & \ding{51}                 & \ding{51}                    \\
    Dual-MRC             & -                       & Span.MRC                    & BERT               & \emph{$\mathcal{D}_{17}$},\emph{$\mathcal{D}_{19}$},\emph{$\mathcal{D}_{20a}$},\emph{$\mathcal{D}_{20b}$}                        & \ding{51}              & -                      & \ding{51}              & \ding{51}               & \ding{51}                & \ding{51}                 & \ding{51}                    \\
    \bottomrule 

    \xrowht{8pt}Ours & \xrowht{8pt}\ding{51}   & \xrowht{8pt}Span.Generation  & \xrowht{8pt}BART  & \emph{$\mathcal{D}_{17}$},\emph{$\mathcal{D}_{19}$},\emph{$\mathcal{D}_{20a}$},\emph{$\mathcal{D}_{20b}$}   & \xrowht{8pt} \ding{51} & \xrowht{8pt} \ding{51} & \xrowht{8pt} \ding{51} & \xrowht{8pt} \ding{51}  & \xrowht{8pt} \ding{51}   & \xrowht{8pt} \ding{51}    & \xrowht{8pt}\ding{51}        \\
    \bottomrule
  \end{tabular}
  \caption{The baselines of our experiments. To further demonstrate that our proposed method is a real unified end-to-end ABSA framework, we present our work in the last row. ``E2E'' is short for End-to-End, which means the model should output all the subtasks' results synchronously rather than requiring any preconditions, e.g., pipeline methods. The ``Datasets''  column refers to the datasets that this baseline is conducted.  }
  \label{tb:baseline}
\end{table*}

\subsection{Datasets}
We evaluate our method on four ABSA datasets. All of them are originated from the Semeval Challenges~\citep{DBLP:conf/semeval/PontikiGPPAM14,DBLP:conf/semeval/PontikiGPMA15,DBLP:conf/semeval/PontikiGPAMAAZQ16}, where only the aspect terms and their sentiment polarities are labeled.

The first dataset(\emph{$\mathcal{D}_{17}$}\footnote{Each dataset only contains a subset of all ABSA subtasks. We use the published year of the dataset to distinguish them.}) is annotated by \citet{DBLP:conf/aaai/WangPDX17}, where the unpaire opinion terms are labeled. The second dataset(\emph{$\mathcal{D}_{19}$}) is annotated by \citet{DBLP:conf/naacl/FanWDHC19}, where they pair  opinion terms with corresponding aspects. The third dataset(\emph{$\mathcal{D}_{20a}$}) is from \citet{DBLP:conf/aaai/PengXBHLS20}. They refine the data in  \textless{}$a$, $o$, $s$\textgreater{} triplet form. The fourth dataset(\emph{$\mathcal{D}_{20b}$}) from \citet{DBLP:conf/emnlp/XuLLB20} is the revised variant of \citet{DBLP:conf/aaai/PengXBHLS20}, where the missing triplets with overlapping opinions are  corrected.  We present the statistics for these four datasets in  Table \ref{tb:data}.

\subsection{Baselines}
To have a fair comparison, we summarize top-performing baselines of all ABSA subtasks. Given different ABSA subtasks, datasets, and experimental setups, existing baselines can be separated into three groups roughly as shown in Table~\ref{tb:baseline}.

The baselines in the first group are conducted on \emph{$\mathcal{D}_{17}$} dataset, covering the \emph{AE}, \emph{OE}, \emph{ALSC}, and \emph{AESC}  subtasks. Span-based method SPAN-BERT~\citep{DBLP:conf/acl/HuPHLL19} and sequence tagging method, IMN-BERT~\citep{DBLP:conf/acl/HeLND19} and RACL-BERT~\citep{DBLP:conf/acl/ChenQ20}, are selected. Specifically, the IMN-BERT model is  reproduced by \citet{DBLP:conf/acl/ChenQ20}. All these baselines are implemented on BERT-Large.

The baselines of the second group are conducted on \emph{$\mathcal{D}_{19}$} dataset, mainly focusing on \emph{AOE}  subtask. Interestingly, we find that sequence tagging method is the main solution for this subtask~\citep{DBLP:conf/naacl/FanWDHC19,DBLP:conf/aaai/WuZDHC20,DBLP:conf/emnlp/VeysehNDDN20}.

The baselines of the third group are mainly conducted on \emph{$\mathcal{D}_{20a}$} and \emph{$\mathcal{D}_{20b}$} datasets, which could  cover almost all the  ABSA subtasks except for one certain subtask depending on the baseline structures.   For the following baselines: RINANTE~\citep{DBLP:conf/acl/DaiS19}, CMLA~\citep{DBLP:conf/aaai/WangPDX17}, Li-unified~\citep{DBLP:conf/aaai/LiBLL19}, the suffix ``+'' in  Table~\ref{tb:baseline}  denotes the corresponding model variant  modified by \citet{DBLP:conf/aaai/PengXBHLS20} for being capable of \emph{AESC},  \emph{Pair}  and \emph{Triplet}.

\subsection{Implement Details}
Following previous studies, we use different metrics according to different subtasks and datasets. Specifically, for the single output subtasks \emph{AE}, \emph{OE}, and \emph{AOE}, the prediction span would be considered  as correct only if it exactly matches the start and the end boundaries. For the ALSC subtask, we require the generated sentiment polarity of the given aspect should be the same as the ground truth. As for compound output subtasks, \emph{AESC}, \emph{Pair} and \emph{Triplet}, a prediction result is correct only when  all the span boundaries and the generated sentiment polarity are  accurately identified. We report the precision (P), recall (R), and F1 scores for all experiments\footnote{Due to the limited space, we would present detailed experiments for each dataset in the Supplementary Material.}.
\begin{table*}[h]
  \centering\small
  \setlength{\tabcolsep}{0.75pt}
  \renewcommand{\arraystretch}{1.3}
  \begin{tabular}{m{2.5cm}m{1.065cm}<{\centering}m{1.065cm}<{\centering}m{1.065cm}<{\centering}m{1.065cm}<{\centering}|m{1.065cm}<{\centering}m{1.065cm}<{\centering}m{1.065cm}<{\centering}m{1.065cm}<{\centering}|m{1.065cm}<{\centering}m{1.065cm}<{\centering}m{1.065cm}<{\centering}m{1.065cm}<{\centering}m{1.065cm}<{\centering}}
    \Xhline{0.08em}
  \multirow{2}{*}{Model} & \multicolumn{4}{c|}{14res}                                        & \multicolumn{4}{c|}{14lap}                                      & \multicolumn{4}{c}{15res}                                        \\
    \cline{2-13}
    & \emph{AE}            & \emph{OE}             & \emph{ALSC}             & \emph{AESC}            & \emph{AE}             & \emph{OE}          & \emph{ALSC}             & \emph{AESC}            & \emph{AE}             & \emph{OE}             & \emph{ALSC}             & \emph{AESC}           \\
    \cline{1-13}
    SPAN-BERT              & 86.71         & -              & 71.75          & 73.68          & 82.34          & -           & 62.5           & 61.25          & 74.63          & -              & 50.28          & 62.29         \\
  IMN-BERT               & 84.06         & 85.10          & 75.67          & 70.72          & 77.55          & \textbf{81.0} & 75.56          & 61.73          & 69.90          & 73.29          & 70.10          & 60.22         \\
  RACL-BERT              & 86.38         & 87.18          & 81.61          & 75.42          & 81.79          & 79.72       & 73.91          & 63.40          & 73.99          & 76.0           & \textbf{74.91} & 66.05         \\
  Dual-MRC               & 86.60         & -              & \textbf{82.04} & \textbf{75.95} & 82.51          & -           & 75.97          & 65.94          & 75.08          & -              & 73.59          & 65.08         \\
    \Xhline{0.05em}
    Ours                   & \textbf{87.07} & \textbf{87.29} & 75.56          & 73.56          & \textbf{83.52} & 77.86       & \textbf{76.76} & \textbf{67.37} & \textbf{75.48} & \textbf{76.49} & 73.91          & \textbf{66.61} \\
    \Xhline{0.08em}
  \end{tabular}
  \caption{Comparison F1 scores for  \emph{AE}, \emph{OE}, \emph{SC}, and \emph{AESC}   on the  $\mathcal{D}_{17}$ dataset~\citep{DBLP:conf/aaai/WangPDX17}. The baseline results are retrieved from \citet{DBLP:journals/corr/abs-2101-00816}. We highlight the best results in bold. It is worth noting that all the baseline results are obtained via BERT-Large, while our results are obtained via BART-Base.}
  \label{tb:wang}
  \end{table*}

\begin{table*}[ht]
  \centering\small
  \setlength{\tabcolsep}{0.75pt} 
  \renewcommand{\arraystretch}{1.3}
  \begin{tabular}{m{2.5cm}m{1.065cm}<{\centering}m{1.065cm}<{\centering}m{1.065cm}<{\centering}|m{1.065cm}<{\centering}m{1.065cm}<{\centering}m{1.065cm}<{\centering}|m{1.065cm}<{\centering}m{1.065cm}<{\centering}m{1.065cm}<{\centering}|m{1.065cm}<{\centering}m{1.065cm}<{\centering}m{1.065cm}<{\centering}m{1.065cm}<{\centering}}
    \Xhline{0.08em}
    \multirow{2}{*}{Model} & \multicolumn{3}{c|}{14res} & \multicolumn{3}{c|}{14lap} & \multicolumn{3}{c|}{15res} & \multicolumn{3}{c}{16res} \\
    \cline{2-13}
    & P       & R      & F1     & P       & R      & F1     & P       & R      & F1     & P       & R      & F1     \\
    \cline{1-13}
    IOG                    & 82.38   & 78.25  & 80.23  & 73.43   & 68.74  & 70.99  & 72.19   & 71.76  & 71.91  & 84.36   & 79.08  & 81.60   \\
  LOTN                   & 84.0      & 80.52  & 82.21  & 77.08   & 67.62  & 72.02  & 76.61   & 70.29  & 73.29  & 86.57   & 80.89  & 83.62  \\
  ONG                    & 83.23   & 81.46  & 82.33  & 73.87   & 77.78  & 75.77  & 76.63   & \textbf{81.14}  & 78.81  & 87.72   & 84.38  & 86.01  \\
  Dual-MRC               & \textbf{89.79}   & 78.43  & 83.73  & 78.21   & \textbf{81.66}  & 79.90   & 77.19   & 71.98  & 74.50   & 86.07   & 80.77  & 83.33  \\
    \Xhline{0.05em}
    Ours                   & 86.01    & \textbf{84.76}  & \textbf{85.38}  & \textbf{83.11}   & 78.13  & \textbf{80.55}  & \textbf{80.12}   & 80.93  & \textbf{80.52}  & \textbf{89.22}   & \textbf{86.67}   & \textbf{87.92}  \\
  \Xhline{0.08em}
  \end{tabular}
  \caption{Comparison results for \emph{AOE}  on the  $\mathcal{D}_{19}$ dataset~\citep{DBLP:conf/naacl/FanWDHC19}. Baselines are from  the original papers. We highlight the best results in bold.}
  \label{tb:fan}
  \end{table*}

  \begin{table*}[!ht]
    \centering\small
    \setlength{\tabcolsep}{0.75pt} 
    \renewcommand{\arraystretch}{1.3}
    \begin{tabular}{m{2.5cm}m{1.065cm}<{\centering}m{1.065cm}<{\centering}m{1.065cm}<{\centering}|m{1.065cm}<{\centering}m{1.065cm}<{\centering}m{1.065cm}<{\centering}|m{1.065cm}<{\centering}m{1.065cm}<{\centering}m{1.065cm}<{\centering}|m{1.065cm}<{\centering}m{1.065cm}<{\centering}m{1.065cm}<{\centering}m{1.065cm}<{\centering}}
      \Xhline{0.08em}
      \multirow{2}{*}{Model}   & \multicolumn{3}{c|}{14res}                     & \multicolumn{3}{c|}{14lap}                      & \multicolumn{3}{c|}{15res}                         & \multicolumn{3}{c}{16res}                                                                                                                                                                                                                                                                                                                                                                                                                                            \\
      \cline{2-13}
                               & \multicolumn{1}{m{1.065cm}<{\centering}}{\emph{AESC}} & \multicolumn{1}{m{1.065cm}<{\centering}}{ \emph{Pair} } & \multicolumn{1}{m{1.065cm}<{\centering}|}{\emph{Triple.}} & \multicolumn{1}{m{1.065cm}<{\centering}}{\emph{AESC}} & \multicolumn{1}{m{1.065cm}<{\centering}}{ \emph{Pair} } & \multicolumn{1}{m{1.065cm}<{\centering}|}{\emph{Triple.}} & \multicolumn{1}{m{1.065cm}<{\centering}}{\emph{AESC}} & \multicolumn{1}{m{1.065cm}<{\centering}}{ \emph{Pair} } & \multicolumn{1}{m{1.065cm}<{\centering}|}{\emph{Triple.}} & \multicolumn{1}{m{1.065cm}<{\centering}}{\emph{AESC}} & \multicolumn{1}{m{1.065cm}<{\centering}}{ \emph{Pair} } & \multicolumn{1}{m{1.065cm}<{\centering}}{\emph{Triple.}} \\
      \hline
      CMLA+ $\dagger$          & 70.62                                          & 48.95                                           & 43.12                                              & 56.90                                          & 44.10                                           & 32.90                                               & 53.60                                          & 44.60                                           & 35.90                                               & 61.20                                          & 50.00                                           & 41.60                                              \\
      RINANTE+ $\dagger$       & 48.15                                          & 46.29                                           & 34.03                                              & 36.70                                          & 29.70                                           & 20.0                                               & 41.30                                          & 35.40                                           & 28.0                                               & 42.10                                          & 30.70                                           & 23.30                                             \\
      Li-unified+ $\dagger$    & 73.79                                          & 55.34                                           & 51.68                                              & 63.38                                          & 52.56                                           & 42.47                                              & 64.95                                          & 56.85                                           & 46.69                                              & 70.20                                          & 53.75                                           & 44.51                                             \\
      Peng-two-stage $\dagger$ & 74.19                                          & 56.10                                            & 51.89                                              & 62.34                                          & 53.85                                           & 43.50                                              & 65.79                                          & 56.23                                           & 46.79                                              & 71.73                                          & 60.04                                           & 53.62                                             \\
      JET-BERT $\sharp$        & -                                              & -                                               & 63.92                                              & -                                              & -                                               & 50.0                                               & -                                              & -                                               & 54.67                                              & -                                              & -                                               & 62.98                                             \\
      Dual-MRC$\dagger$        & 76.57                                          & 74.93                                           & 70.32                                              & 64.59                                          & 63.37                                           & 55.58                                              & 65.14                                          & 64.97                                           & 57.21                                              & 70.84                                          & 75.71                                           & 67.40                                             \\
      \Xhline{0.05em}
      Ours                     & \textbf{78.47}                                 & \textbf{77.68}                                  & \textbf{72.46}                                     & \textbf{68.17}                                 & \textbf{66.11}                                  & \textbf{57.59}                                     & \textbf{69.95}                                 & \textbf{67.98}                                  & \textbf{60.11}                                     & \textbf{75.69}                                 & \textbf{77.38}                                  & \textbf{69.98}                                    \\

      \Xhline{0.09em}
    \end{tabular}
    \caption{Comparison F1 scores for  \emph{AESC}, \emph{Pair} and \emph{Triplet} on the  $\mathcal{D}_{20a}$ dataset~\citep{DBLP:conf/aaai/PengXBHLS20}. The baseline results with ``$\dagger$'' are retrieved from \citet{DBLP:journals/corr/abs-2101-00816}, and result with ``$\sharp$'' is from \citet{DBLP:conf/emnlp/XuLLB20}. We highlight the best results in bold.}
    \label{tb:penga}
  \end{table*}

  \begin{table*}[!ht]
    \centering\small
    \setlength{\tabcolsep}{0.75pt} 
    \renewcommand{\arraystretch}{1.3}
    \begin{tabular}{m{2.5cm}m{1.065cm}<{\centering}m{1.065cm}<{\centering}m{1.065cm}<{\centering}|m{1.065cm}<{\centering}m{1.065cm}<{\centering}m{1.065cm}<{\centering}|m{1.065cm}<{\centering}m{1.065cm}<{\centering}m{1.065cm}<{\centering}|m{1.065cm}<{\centering}m{1.065cm}<{\centering}m{1.065cm}<{\centering}m{1.065cm}<{\centering}}
      \Xhline{0.08em}
      \multirow{2}{*}{Model} & \multicolumn{3}{c|}{14res} & \multicolumn{3}{c|}{14lap} & \multicolumn{3}{c|}{15res} & \multicolumn{3}{c}{16res}                                                                                            \\
      \cline{2-13}
                             & P                          & R                          & F1                         & P                         & R     & F1             & P     & R     & F1             & P     & R     & F1             \\
      \cline{1-13}
      CMLA+                  & 39.18                      & 47.13                      & 42.79                      & 30.09                     & 36.92 & 33.16          & 34.56 & 39.84 & 37.01          & 41.34 & 42.1  & 41.72          \\
      RINANTE+               & 31.42                      & 39.38                      & 34.95                      & 21.71                     & 18.66 & 20.07          & 29.88 & 30.06 & 29.97          & 25.68 & 22.3  & 23.87          \\
      Li-unified+            & 41.04                      & \textbf{67.35}                      & 51.0                       & 40.56                     & 44.28 & 42.34          & 44.72 & 51.39 & 47.82          & 37.33 & 54.51 & 44.31          \\
      Peng-two-stage         & 43.24                      & 63.66                      & 51.46                      & 37.38                     & 50.38 & 42.87          & 48.07 & 57.51 & 52.32          & 46.96 & 64.24 & 54.21          \\
      JET-BERT               & \textbf{70.56}                      & 55.94                      & 62.40                       & 55.39                     & 47.33 & 51.04          & \textbf{64.45} & 51.96 & 57.53          & \textbf{70.42} & 58.37 & 63.83          \\
      \Xhline{0.05em}
      Ours                   & 65.52                     & 64.99                      & \textbf{65.25}             & \textbf{61.41}                     & \textbf{56.19} & \textbf{58.69} & 59.14 & \textbf{59.38} & \textbf{59.26} & 66.6 & \textbf{68.68}& \textbf{67.62} \\
      \Xhline{0.08em}
    \end{tabular}
    \caption{Comparison results for \emph{Triplet} on the  $\mathcal{D}_{20b}$ dataset~\citep{DBLP:conf/emnlp/XuLLB20}. Baselines are from \citep{DBLP:conf/emnlp/XuLLB20}. We highlight the best results in bold.}
    \label{tb:pengb}
  \end{table*}

\subsection{Main Results}
On   $\mathcal{D}_{17}$ dataset~\citep{DBLP:conf/aaai/WangPDX17}, we compare our method for \emph{AE}, \emph{OE}, \emph{ALSC}, and \emph{AESC}. The comparison results are shown in Table~\ref{tb:wang}.  Most of our results achieve better or comparable results to baselines.  However, these baselines yield competitive results based on the BERT-Large pre-trained models. While our results are achieved on the BART-Base model with almost half parameters. This shows that our framework is more suitable for these ABSA subtasks.

On $\mathcal{D}_{19}$ dataset~\citep{DBLP:conf/naacl/FanWDHC19}, we compare our method for \emph{AOE}. The comparison results are shown in Table~\ref{tb:fan}.
We can observe that  our method achieves  significant P/R/F1 improvements on 14res, 15res, and 16res. Additionally, we notice that  our F1 score  on 14lap is close to the previous SOTA result. This is probably caused by the dataset domain difference as the 14lap is the laptop comments while the others are restaurant comments.

On $\mathcal{D}_{20a}$ dataset~\citep{DBLP:conf/aaai/PengXBHLS20}, we compare our method for  \emph{AESC}, \emph{Pair}, and \emph{Triplet}. The comparison results are shown in Table~\ref{tb:penga}. We can observe that our proposed method is able to outperform  other baselines on all datasets. Specifically, we achieve the better results  for \emph{Triplet}, which  demonstrates the effectiveness of our method on capturing interactions among aspect terms, opinion terms, and sentiment polarities. We also observe that the Span-based methods show superior performance to sequence tagging methods. This may be caused by the higher compositionality of candidate labels in  sequence tagging methods~\citep{DBLP:conf/acl/HuPHLL19}. As the previous SOTA method, the Dual-MRC shows competitive performance by utilizing  the span-based extraction method and the MRC mechanism. However, their inference process is  not an  end-to-end process.

On $\mathcal{D}_{20b}$ dataset~\citep{DBLP:conf/emnlp/XuLLB20}, we compare our method for \emph{Triplet}. The comparison results can be found in Table~\ref{tb:pengb}. Our method achieves the best results with nearly 7 F1 points improvements on 14res, 15res, and 16res. Our method achieves nearly 13, 9, 7, 12 points improvements on each dataset for the recall scores compared with other baselines. This also explains the drop performance of the precision score. Since $\mathcal{D}_{20b}$ is  refined from $\mathcal{D}_{20a}$, we specifically compare the \emph{Triplet} results of the corresponding dataset in $\mathcal{D}_{20a}$ and $\mathcal{D}_{20b}$. Interestingly, we discover that all baselines have a much bigger performance change on 15res. We conjecture the distribution differences may be the cause reason. In conclusion, all the experiment results confirm that our proposed method, which unifies the training and the inference to an  end-to-end generative framework, provides a new SOTA solution for the whole ABSA task.


\section{Framework Analysis}
\label{sec:ana}
To better understand our proposed framework, we conduct analysis experiments on the $\mathcal{D}_{20b}$ dataset~\citep{DBLP:conf/emnlp/XuLLB20}.

To validate whether our proposed framework could adapt to the generative ABSA task, we metric the invalid predictions for the \emph{Triplet}. Specifically, since the \emph{Triplet} requires the prediction format like $[a^{s}, a^{e}, o^{s}, o^{e}, s^p]$, it is mandatory that one valid triplet prediction should be in length 5, noted as ``5-len'', and obviously all end index should be larger than the corresponding start index, noted as ``ordered prediction''. We calculate $\frac{  number \  of \  non-5-len}{total\  prediction}$, referred to as the ``Invalid size'', and the $\frac{ number\  of\  non-ordered\  prediction}{total\  5-len\  prediction}$, referred to as the ``Invalid order''.  The ``Invalid token'' means the $a^s$ is not the start of a token, instead, it is the index of an inside subword. From Table \ref{tb:error}, we can observe that BART could learn this task form easily as the low rate for all the three metrics, which demonstrate that the generative framework for ABSA is not only a theoretically unified task form but also  a realizable framework in practical. We remove these invalid predictions in our implementation of experiments.

\begin{figure}[ht]
  \centering
  \hspace{-0.5cm}
  \begin{subfigure}{0.5\textwidth}
  \centering
  \begin{tikzpicture}[scale=0.4]
    \begin{axis}[
      xtick ={1,2,3,4},
      xlabel=Beam Size,
      ylabel=F1 score,
      ytick distance = 3,
      grid=major,
      title=Pair Extraction F1 scores
    ]
    \addplot coordinates{
      (1, 63.61)
      (2, 64.52)
      (4, 64.22)
    };

    \addplot coordinates{
      (1, 69.59)
      (2, 69.99)
      (4, 70.05)
    };

    \addplot coordinates{
      (1, 73.58)
      (2, 73.58)
      (4, 73.58)
    };
    \addplot coordinates{
      (1, 76.02)
      (2, 76.27)
      (4, 76.27)
    };
    \end{axis}
  \end{tikzpicture}
  \end{subfigure}
  \begin{subfigure}{0.5\textwidth}
    \centering
    \begin{tikzpicture}[scale=0.4]
      \begin{axis}[ legend style={at={(1.1,0.5)},anchor=north,cells={align=left}},
        xlabel=Beam Size,
        ylabel=F1 score,
        xtick ={1,2,3,4},
        ytick distance = 3,
        legend columns=1,
        grid=major,
        title=Triplet Extraction F1 scores
      ]
      \addplot coordinates {
        (1,57.4)
        (2,58.06)
        (4,58.06)
      };
      \addplot coordinates {
        (1,61.06)
        (2,61.56)
        (4,61.62)
      };
      \addplot coordinates {
        (1,67.89)
        (2,67.89)
        (4,67.89)
      };
      \addplot coordinates {
        (1,70.47)
        (2,70.45)
        (4,70.45)
      };
      \legend{14lap, 14res, 15res, 16res}
      \end{axis}
    \end{tikzpicture}
    \end{subfigure}


  \caption{The F1 change curve with the increment of beam size on the dev set of \emph{$\mathcal{D}_{20b}$}. The beam size seems to have little effect on the F1 scores. }\label{fig:beam}
\end{figure}
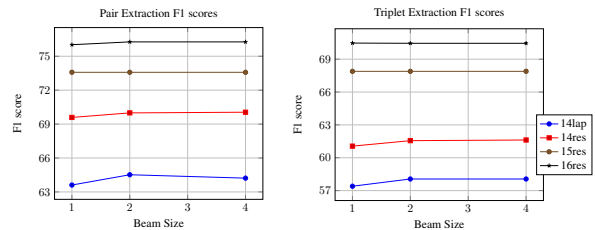

As shown in Table~\ref{fig:beam}, we give some analysis on the impact of the beam size, as we are a generation method. However, the beam size seems to have little impact on the F1 scores.

\begin{table}[h]
  \centering
  \setlength{\tabcolsep}{1pt} 
  \renewcommand{\arraystretch}{1.3}
  \begin{tabular}{m{2.5cm}m{1.15cm}<{\centering}m{1.15cm}<{\centering}m{1.15cm}<{\centering}m{1.15cm}<{\centering}}
    \Xhline{0.08em}
  Errors               & 14res  & 14lap  & 15res  & 16res  \\
  \cline{1-5}
  Invalid size         & 0.48\% & 0.77\% & 1.41\% & 1.40\% \\
  Invalid order        & 1.75\% & 3.70\% & 3.26\% & 3.26\% \\
  Invalid token      & 0.48\% & 0.78\% & 1.02\% & 1.02\% \\
  \Xhline{0.08em}
  \end{tabular}
  \caption{The errors  for \emph{Triplet} on the test set of the \emph{$\mathcal{D}_{20b}$}.  }
  \label{tb:error}
  \end{table}

\section{Conclusion}
This paper summarizes the seven ABSA subtasks and previous studies, which shows that there exist divergences on all the input, output, and task type sides. Previous studies have limitations on handling all these divergences in a unified framework. We propose to convert all the ABSA subtasks to a unified generative task. We implement the BART to generate the target sequence in an end-to-end process based on the unified task formulation. We conduct massive experiments on public datasets for seven ABSA subtasks and achieve significant improvements on most datasets. The experimental results demonstrate the effectiveness of our method. Our work leads to several promising directions, such as sequence-to-sequence framework on other tasks, and data augmentation.

\section*{Acknowledgements}
We would like to thank the anonymous reviewers for their insightful comments. The discussion with colleagues in AWS Shanghai AI Lab was quite fruitful. We also thank the developers of fastNLP\footnote{\url{https://github.com/fastnlp/fastNLP}. FastNLP is a natural language processing python package.} and fitlog\footnote{\url{https://github.com/fastnlp/fitlog}. Fitlog is an experiment tracking package}. This work was supported by the National Key Research and Development Program of China (No. 2020AAA0106700) and National Natural Science Foundation of China (No. 62022027).

\section*{Ethical Considerations}
For the consideration of ethical concerns, we would make detailed description  as follows:

(1) All the experiments are conducted on existing datasets, which are derived from public scientific papers.

(2) We describe the characteristics of the datasets in a specific section. Our analysis is consistent with the results.

(3) Our work does not contain identity characteristics. It does not harm anyone.

(4) Our experiments do not  need a lot of computer resources compared to pre-trained models.

(5) We will open source all our code.

\bibliographystyle{acl_natbib}
\bibliography{anthology,acl2021}


\clearpage
\appendix
\section{Supplemental Material}

\subsection{Experimental Environment} \label{supply:random_search}

We use the triangular learning rate warmup. All experiments are conducted in the Nvidia Ge-Force RTX-3090 Graphical Card with 24G graphical memory. 

The averages running time for experiments on each dataset is less than 15 minutes. The number of parameters is as follows:

$\bullet$ BART-Base model: 12 layers, 768 hidden dimensions and 16 heads with the total number of parameters, 139M;

$\bullet$ BERT-Base model: 12 layers, 768 hidden dimensions and 12 heads with the total number of parameters, 110M.


\subsection{Decoding Algorithm for Different Datasets}
In this part, we introduce the decoding algorithm we used to convert the predicted target sequence $Y$ into the target span set $L$.  These algorithm can be found in Algorithm \ref{al4}, \ref{al2}, \ref{al3}.

\begin{algorithm}[!h]
  \begin{algorithmic}[1]
    \caption{Decoding Algorithm for the \emph{AOE} subtask} \label{al4}
    \Require Number of tokens in the input sentence $n$, target sequence $Y=[y_1, ..., y_m]$ and $y_i \in [1, n+|C|]$, $L_T$ is a given length for different tasks.
    \Ensure  Target span set $L=\{(o_1^s, o_1^e, ..., o_i^s, o_i^e)\}$
    \State $L=\{\}, e=[], i=3$
    \While{$i<=m$}
      \State $y_i = Y[i]$
      \State $e.append(y_i)$
      \State $i+=1$
    \EndWhile
    \State $L.add(e)$
    \State \Return{$L$}
  \end{algorithmic}  
\end{algorithm}

\begin{algorithm}[!h]
  \begin{algorithmic}[1]
    \caption{Decoding Algorithm for the \emph{AESC} Subtask} \label{al2}
    \Require Number of tokens in the input sentence $n$, target sequence $Y=[y_1, ..., y_m]$ and $y_i \in [1, n+|C|]$
    \Ensure  Target span set $L=\{(a_1^s, a_1^e, s_1), ..., (a_i^s, a_i^e, s_i)\}$
    \State $L=\{\}, e=[], i=1$ 
    \While{$i<=m$}
      \State $y_i = Y[i]$
      \If{$y_i>n$}
      \State  $L.add((e, C_{y_i-n}))$
      \State $e=[]$
      \Else
      \State $e.append(y_i)$
      \EndIf
      \State $i+=1$
    \EndWhile
    \State \Return{$L$}
  \end{algorithmic}  
\end{algorithm}

\begin{algorithm}[!h]
    \begin{algorithmic}[1]
      \caption{Decoding Algorithm for the \emph{AE}/\emph{OE}/\emph{Pair} subtasks} \label{al3}
      \Require Number of tokens in the input sentence $n$, target sequence $Y=[y_1, ..., y_m]$ and $y_i \in [1, n+|C|]$, $L_T$ is a given length for different tasks.
      \Ensure  Target span set $L=\{x_1, ..., x_i\}$($x_i$ is $(a_i^s, a_i^e)$, $(o_i^s, o_i^e)$ and $(a_i^s, a_i^e, o_i^s, o_i^e)$ for \emph{AE}/\emph{OE}/\emph{Pair}, respectively)
      \State $L=\{\}, e=[], i=1$ 
      \While{$i<=m$}
        \State $y_i = Y[i]$
        \If{$len(e)==L_T$}
        \State  $L.add((e, C_{y_i-n}))$
        \State $e=[]$
        \EndIf
        \State $e.append(y_i)$
        \State $i+=1$
      \EndWhile
      \State \Return{$L$}
    \end{algorithmic}  
\end{algorithm}

\subsection{Detailed Experimental Setup}
\textbf{Experiments on each dataset}

As the different subtasks are conducted on different datasets, specifically, we conduct the following experiments on each dataset:

$\bullet$  On the \emph{$\mathcal{D}_{17}$} dataset, we conduct the AESC and the OE  in multi-task learning method. To that end, we feed the pre-defined task tags ``$<$AESC$>$'' and ``$<$OE$>$'' to the decoder first. For example, for the input ``\emph{The \textbf{drinks} are always \uwave{well made} and \textbf{wine selection} is \uwave{fairly priced}}'' from \emph{$\mathcal{D}_{17}$} dataset, we define the AESC sequence and the OE target sequence as  ``$<$\underline{AESC}$>$, 1, 1, POS, 7, 8, POS, $<$/s$>$'' and ``$<$\underline{OE}$>$, 4, 5, 10, 11, $<$/s$>$''.

$\bullet$  On the \emph{$\mathcal{D}_{19}$} dataset, we conduct the AOE. As the AOE subtask requires to detect the opinion terms given aspect  terms in advance, the aspect terms need to be fed to our decoder first. For the aforementioned example sentence from \emph{$\mathcal{D}_{19}$} dataset, we define the AOE target sequence as `` \underline{1}, \underline{1}, 4, 5, $<$/s$>$'' and the `` \underline{7}, \underline{8}, 10, 11, $<$/s$>$''.

$\bullet$ On the \emph{$\mathcal{D}_{20a}$} and \emph{$\mathcal{D}_{20b}$} datasets, we conduct the Triplet Extraction. For the aforementioned example sentence from \emph{$\mathcal{D}_{20a}$} and \emph{$\mathcal{D}_{20b}$} dataset, we define the Triplet target sequence as ``1, 1, 4, 5, POS, 7, 8, 10, 11, POS, $<$/s$>$''.\newline
\textbf{Specific Subtask Metrics}

$\bullet$ On the \emph{$\mathcal{D}_{17}$} dataset, we get the AESC and OE results directly. Following previous work, we only calculate the metrics for AESC and ALSC from those true positive AE predictions.  Specifically, the F1 

$\bullet$ On the \emph{$\mathcal{D}_{19}$} dataset, we get the AOE results directly. The metrics for AOE are standard Precision, Recall and the F1 score.

$\bullet$ On the \emph{$\mathcal{D}_{20a}$} and \emph{$\mathcal{D}_{20b}$} datasets, we get the Triplet results directly.  We preserve the $<$AT,OT$>$ for Pair metric and $<$AT, SP$>$ for AESC metric. The metrics for them are standard Precision, Recall and the F1 score.

\end{document}